\documentclass[a4paper,11pt]{article}

\usepackage{graphicx}  
\usepackage{url}       
\usepackage{times}
\usepackage{natbib}
\usepackage[margin=25mm]{geometry}

\usepackage{enumerate}
\usepackage{linguex}
\usepackage{xcolor}

\newcommand{\td}[1]{{\bf TODO: #1}}

\newcommand{\notiwcs}[1]{}

\title{Don't Blame Distributional Semantics if it can't do Entailment}
\date{}

\author{Matthijs Westera \hspace{4.5cm}  Gemma Boleda \\
		Universitat Pompeu Fabra, Barcelona, Spain\\
		\texttt{\{firstname.lastname\}@upf.edu}
}

\begin{document}
	
\maketitle

\thispagestyle{empty}
\pagestyle{empty}

\begin{abstract}
	\noindent Distributional semantics has had enormous empirical success in Computational Linguistics and Cognitive Science in modeling various semantic phenomena, such as semantic similarity, and distributional models are widely used in state-of-the-art Natural Language Processing systems. 
	However, the theoretical status of distributional semantics within a broader theory of language and cognition is still unclear: 
	What does distributional semantics model? 
	Can it be, on its own, a fully adequate model of the meanings of linguistic expressions? 
	The standard answer is that distributional semantics is not fully adequate in this regard, because it falls short on some of the central aspects of \emph{formal} semantic approaches: truth conditions, entailment, reference, and certain aspects of compositionality. 
	We argue that this standard answer rests on a misconception: 
	These aspects do not belong in a theory of expression meaning, they are instead aspects of \emph{speaker} meaning, i.e., communicative intentions in a particular context. 
	In a slogan: words do not refer, speakers do. 
	Clearing this up enables us to argue that distributional semantics on its own is an adequate model of expression meaning. 
	Our proposal sheds light on the role of distributional semantics in a broader theory of language and cognition, its relationship to formal semantics, and its place in computational models.
\end{abstract}


\section{Introduction}
\label{sec:intro}

Distributional semantics has emerged as a promising model of certain `conceptual' aspects of linguistic meaning (e.g., \citealt{landauer1997solution, turney2010frequency, baroni2010distributional, lenci2018distributional}) and as an indispensable component of applications in Natural Language Processing (e.g., reference resolution, machine translation, image captioning; especially since \citealt{mikolov2013distributed}).
Yet its theoretical status within a general theory of meaning and of language and cognition more generally is not clear (e.g., \citealt{Lenci:08, erk2010word, BoledaHerbelot:16, lenci2018distributional}).
In particular, it is not clear whether distributional semantics can be understood as an actual model of expression meaning -- what \citet{Lenci:08} calls the `strong' view of distributional semantics -- or merely as a model of something that \emph{correlates} with expression meaning in certain partial ways -- the `weak' view.
In this paper we aim to resolve, in favor of the `strong' view, the question of what exactly distributional semantics models, what its role should be in an overall theory of language and cognition, and how its contribution to state of the art applications can be understood.
We do so in part by clarifying its frequently discussed but still obscure relation to formal semantics.

Our proposal relies crucially on the distinction between what linguistic \emph{expressions} mean outside of any particular context, and what \emph{speakers} mean by them in a particular context of utterance.
Here, we term the former \textbf{expression meaning} and the latter \textbf{speaker meaning}.\footnote{
	English inconveniently conflates what speakers do and what expressions do in a single verb ``to mean''.
	In other languages the two types of `meaning' go by different names, e.g., in Dutch, sentences `betekenen' (mean, lit.\ `be-sign' or `signify') while speakers `bedoelen' (mean, lit.\ `be-goal').}
At least since \citealt{grice1968utterer} this distinction is generally acknowledged to be crucial to account for how humans communicate via language.
Nevertheless, the two notions are sometimes confused, and we will point out a particularly widespread confusion in this paper.
Consider an example, one which will recur throughout this paper:
\ex. \label{ex:red-cat} The red cat is chasing a mouse. 

The expression ``the red cat'' in this sentence can be used to refer to a cat with red hair (which is actually orangish in color) or to a cat painted red; ``a mouse'' to the animal or to the computer device; and in the right sort of context the whole sentence can be used to describe, for instance, a red car driving behind a motorbike.
It is uncontroversial that the same expression can be used to communicate very different speaker meanings in different contexts.
At the same time, it is likewise uncontroversial that not \emph{anything goes}: what a speaker can reasonably mean by an expression in a given context -- with the aim of being understood by an addressee -- is constrained by its (relatively) context-invariant expression meaning.
An important, long-standing question in linguistics and philosophy is what type of object could play the role of expression meaning, i.e., as a context-invariant common denominator of widely varying usages.

There exist two predominant candidates for a model of expression meaning: distributional semantics and formal semantics.
\textbf{Distributional semantics} assigns to each expression, or at least each word, a high-dimensional, numerical vector, one which represents an abstraction over occurrences of the expression in some suitable dataset, i.e., its \emph{distribution} in the dataset.
\textbf{Formal semantics} assigns to each expression, typically via an intermediate, logical language, an interpretation in terms of reference to entities in the world, their properties and relations, and ultimately truth values of whole sentences.\footnote{
	Our formulation covers only the predominant, model-theoretic (or truth-conditional, referential) type of formal semantics, not, e.g., proof-theoretic semantics.
	We concentrate on this for reasons of space, but our proposal applies more generally.
}
To illustrate the two approaches, simplistically (and without intending to commit to any particular formal semantic analysis or (compositional) distributional semantics -- see Section~\ref{sec:compositionality}):
\ex. \label{ex:red-cat-formal} The red cat is chasing a mouse. \\[2pt]
\textbf{Formal semantics:} \ \ \ \ $\iota x (\textsc{Red}(x) \land \textsc{Cat}(x) \land \exists y (\textsc{Mouse}(y) \land \textsc{Chase}(x,y)))$ \\[4pt]
\textbf{Distributional semantics:} \ \ \ $^\nearrow \ \ _\searrow \ \ _\swarrow \ \rightarrow \ \ _\downarrow \ \ ^\nearrow \ \leftarrow$ \ \ \ \ \ \ (i.e., a vector for each word)

Distributional and formal semantics are often regarded as two models of expression meaning that have \emph{complementary strengths and weaknesses} and that, accordingly, must somehow be combined for a more complete model of expression meaning (e.g., \citealt{Beltagy:13montague, Erk:13towards, baroni2014frege, asher2016integrating, BoledaHerbelot:16}).
For instance, in these works the vectors of distributional semantics are regarded as capturing lexical or conceptual aspects of meaning but not, or insufficiently so, truth conditions, reference, entailment and compositionality -- and vice versa for formal semantics.\footnote{\label{fn:onitsown}To clarify: when it is said that distributional semantics falls short, this pertains to distributional semantics on its own, i.e., a set of word vectors, combined perhaps with some basic algebraic operations or, at most, a simple classifier.
	By contrast, when distributional semantics is incorporated in a larger model (see section~\ref{sec:whatwemean}) the resulting system as a whole can be very successful.
}\notiwcs{\textsuperscript{,}\footnote{The failure of distributional semantics to account for entailment (moderate successes on short phrases notwithstanding) has received particular attention in relation to the \emph{Distributional Inclusion Hypothesis} (\citealt{geffet2005distributional}), according to which entailment would correspond to distributional inclusion.}}

\textbf{Contrary to this common perspective, we argue that distributional semantics on its own can in fact be a fully satisfactory model of expression meaning}, i.e., the `strong' view of distributional semantics in \citealt{Lenci:08}.
Crucially, we will do so \emph{not} by trying to show that distributional semantics can do all the things formal semantics does -- we think it clearly cannot, at least not on its own -- but by explaining that a semantics \emph{should not} do all those things.
In fact, formal semantics is mistaken about its job description, a mistake that we trace back, following a long strand in both philosophical and psycho-linguistic literature, to a failure to properly distinguish speaker meaning and expression meaning.
By clearing this up we aim to contribute to a firmer theoretical understanding of distributional semantics, of its role in an overall theory of communication, and of its employment in current models in NLP.

\section{What we mean by distributional semantics}
\label{sec:whatwemean}

By distributional semantics we mean, in this paper, a broad family of models that assign (context-invariant) numerical vector representations to words, which are computed as abstractions over occurrences of words in contexts.
Implementations of distributional semantics vary, primarily, in the notion of context and in the abstraction mechanism used.
A \textbf{context} for a word is typically a text in which it occurs, such as a document, sentence or a set of neighboring words, but it can also contain images (e.g., \citealt{feng2010visual, silberer2017visually}) or audio (e.g., \citealt{lopopolo2015sound}) -- in principle any place where one may encounter a word could be used.
Because of how distributional models work, words that appear in similar contexts end up being assigned similar representations.
At present, all models need large amounts of data to compute high-quality representations.
The closer these data resemble our experience as language learners, the more distributional semantics is expected to be able in principle to generate accurate representations of -- as we will argue -- expression meaning.

As for the \textbf{abstraction mechanism} used, \cite{baroni2014don} distinguish between classic 
``count-based'' methods, which work with co-occurrence statistics between words and contexts,
and ``prediction-based'' methods, which instead apply machine learning techniques (artificial neural networks) to induce representations based on a prediction task, typically predicting the context given a word.
For instance, the Skip-Gram model of \cite{mikolov2013linguistic} would, applied to example~\ref{ex:red-cat}, try to predict the words ``the'', ``red'', ``is'', ``chasing'', etc.\ from the presence of the word ``cat'' (more precisely, it would try to make these context words more likely than randomly sampled words, like ``democracy'' or ``smear'').
By training a neural network on such a task, over a large number of words in context, the first layer of the network comes to represent words as vectors, usually called \emph{word embeddings} in the neural network literature.
These word embeddings contain information about the words that the network has found useful for the prediction task.

In both count-based and prediction-based methods, the resulting vector representations encode abstractions over the distributions of words in the dataset, with the crucial property that words that appear in similar contexts are assigned similar vector representations.\footnote{
	Both methods also share the characteristic that the dimensions of the high-dimensional space are automatically induced, and hence not directly interpretable (this is the main way in which they are different from traditional semantic features; see \citealt{boleda2015distributional}).
	As a consequence, much work exploring distributional semantic models has relied not on the dimensions themselves but on geometric relations between words, in particular the notion of similarity (e.g., measured by cosine; as an anonymous reviewer notes, such technical notions of similarity need not completely align with semantic similarity in a more intuitive sense).
}
Our arguments in this paper apply to both kinds of methods for distributional semantics.

Word embeddings emerge not just from models that are expressly designed to yield word representations (such as \citealt{mikolov2013linguistic}).
Rather, any neural network model that takes words as input, trained on whatever task, must `embed' these words in order to process them -- hence any such model will result in word embeddings (e.g., \citealt{collobert2008unified}).
Neural network models for language are trained for instance on language modeling (e.g., word prediction;~\citealt{mikolov2010recurrent,Peters:2018elmo}) or Machine Translation~\citep{bahdanau+15}.
As long as the data on which these models are trained consist of word-context pairs, the resulting word embeddings qualify, for present purposes, as implementations of distributional semantics, and our proposal in the current paper applies also to them.
Of course some implementations within this broad family may be better than others, and the type of task used is one parameter to be explored: It is expected that the more the task requires a human-like understanding of language, the better the resulting word embeddings will represent -- as we will argue -- the meanings of words.
But our arguments concern the theoretical underpinnings of the distributional semantics framework more broadly rather than specific instantiations of it.

Lastly, some implementations of distributional semantics impose biases, during training, for obtaining word vectors that are more useful for a given task.
For instance, to obtain word vectors useful for predicting lexical entailment (e.g., that being a cat entails being an animal), \citet{vulic2017specialising} impose a bias for keeping the vectors of supposed hypernyms, like ``cat'' and ``animal'', close together (more precisely: in the same direction from the origin but with different magnitudes).
This kind of approach presupposes, incorrectly as we will argue, that distributional semantics \emph{should} account for entailment.
It results in word vectors that are more useful for a particular task, but the model will be worse as a model of expression meaning.
We will return to this type of approach in section~\ref{sec:concepts}.

\section{Distributional semantics as a model of expression meaning}
\label{sec:ds}

We present two theoretical reasons why distributional semantics is attractive as a model of expression meaning, before arguing in section~\ref{sec:reference} that it can also be sufficient.
\notiwcs{
The first reason is that abstractions over past usage, as generated by distributional semantics, are a minimal and explanatory, hence theoretically parsimonious candidate for expression meaning.
The second reason is that the representations generated by distributional semantics can be regarded as concepts, more precisely concepts of words, which are a necessary starting point for interpretation.
}

\subsection{Reason 1: Meaning from use; abstraction and parsimony}
\label{sec:meaningfromuse}

We take it to be uncontroversial that what expressions mean is to be explained at least in part in terms of how they are used by speakers of the relevant linguistic community (e.g., \citealt{Wittgenstein:53, grice1968utterer}).\footnote{
	For compatibility with a more cognitive, single-agent perspective of language, such as \emph{I-language} in the work of Chomsky (e.g., \citeyear{Chomsky:86}), this could be restricted to the uses of a word as experienced by a single agent when learning the language.}
A similar view has motivated work on distributional semantics (e.g., \citealt{Lenci:08}; also at its conception, e.g., \citealt{harris1954distributional}).
For instance, what the word ``cat'' means is to be explained at least in part in terms of the fact that speakers have used it to refer to cats, to describe things that resemble cats, to insult people in certain ways, and so on.
Note that the usages of words generally resist systematic categorization into definable senses, and attempts to characterize word meaning by sense enumeration generally fail (e.g., \citealt{kilgarriff1997don, hanks2000word, erk2010word}; cf.\ \citealt{pustejovsky1995}).

A minimal, parsimonious way of explaining the meaning of an expression in terms of its uses is to say simply that \textbf{the meaning of an expression is an abstraction over its uses}.
Such abstractions are, of course, exactly what distributional semantics delivers, and the view that it corresponds to expression meaning is what \citet{Lenci:08} calls the `strong' view of distributional semantics.
Distributional semantics is especially parsimonious because it relies on (mostly) domain-independent mechanisms for abstraction (e.g., principal components analysis; neural networks).
Of course not all implementations are equally adequate, or equally parsimonious; there are considerable differences both in the abstraction mechanism relied upon and in the dataset used (see section~\ref{sec:whatwemean}).
But the family as a whole, defined by the core tenet of associating with each word an abstraction over its use, is highly suitable in principle for modeling expression meaning.
This makes the `strong' view of distributional semantics attractive.

An alternative to the `strong' view is what \citet{Lenci:08} calls the `weak' view: that an abstraction over use may be \emph{part} of what determines expression meaning, but that more is needed.
This view underlies for instance the common assumption that a more complete model of expression meaning would require integrating distributional and formal semantics (e.g., \citealt{Beltagy:13montague, Erk:13towards, baroni2014frege, asher2016integrating, BoledaHerbelot:16}).
But in section~\ref{sec:reference} we argue that the notions of formal semantic, like reference, truth conditions and entailment, do not belong at the level of expression meaning in the first place, and, accordingly, that distributional semantics can be sufficient as a model of expression meaning.
Theoretical parsimony dictates that we opt for the least presumptive approach compatible with the empirical facts, i.e., with what a theory of expression meaning should account for.\notiwcs{\footnote{
	Besides this methodological argument for a parsimonious notion of expression meaning, a cognitive/evolutionary argument can also be made; see, e.g., \citealt{brochhagen2018coevolution}.}}

Some authors equate the meaning of an expression not with an abstraction over all uses, but only \emph{stereotypical} uses: what an expression means would be what a stereotypical speaker in a stereotypical context means by it (e.g., \citealt{schiffer1972meaning, bennett1976linguistic, soames2002beyond}).
This approach is appealing because it does justice to native speaker's intuitions about expression meaning, which are known to reflect stereotypical speaker meaning (see Section~\ref{sec:reference}).
However, several authors have pointed out that stereotypical speaker meaning is ultimately not an adequate notion of expression meaning (e.g., \citealt{Bach:02, Recanati:04}).
To see just one reason why, consider the following arbitrary example:
\ex. \label{ex:marry1} Jack and Jill got married.

A stereotypical use of this expression would convey the speaker meaning that Jack and Jill got married \emph{to each other}.
But this cannot be the (context-invariant) meaning of the expression ``Jack and Jill got married'', or else the following additions would be redundant and contradictory, respectively:\footnote{
	An anonymous reviewer rightly points out that this presupposes that notions like redundancy and contradiction apply to expression meanings. We think they don't (see Section~\ref{sec:reference}), at least not in their strictly logical senses, but they would if expression meaning were to be construed as stereotypical speaker meaning, which is the position we are criticizing here.}
\ex. \label{ex:marry2} Jack and Jill got married \emph{to each other}.

\ex. \label{ex:marry3} Jack and Jill got married \emph{to their respective childhood friends}.

Hence the stereotypical speaker meaning of \ref{ex:marry1} cannot be its expression meaning.
For many more examples and discussion see \citealt{Bach:02}.
Another challenge for defining expression meaning as stereotypical speaker meaning is that of having to define ``stereotypical''.
It cannot be defined simply as the most frequent type, because that presupposes that uses can be categorized into clearly delineated, countable types.
Moreover, an `empty' context is a context too, and not the most stereotypical one.

Summing up: what an expression means depends on how speakers use it, but the uses of an expression more generally resist systematic categorization into enumerable senses, and selecting a stereotypical use isn't adequate either. 
Equating expression meaning with an abstraction over all uses, as the `strong' view of distributional semantics has it, is more adequate, and particularly attractive for reasons of parsimony.

\subsection{Reason 2: Distributional semantics as a model of concepts}
\label{sec:concepts}

Another reason why distributional semantics is attractive as a model of expression meaning is the following.
As mentioned in section~\ref{sec:intro}, distributional semantics is often regarded as a model of `conceptual' aspects of meaning (e.g., \citealt{landauer1997solution, baroni2010distributional, BoledaHerbelot:16}).
This view seems to be motivated in part empirically: distributional semantics is successful at what are intuitively conceptual tasks, like modeling word similarity, priming and analogy.
Moreover, it aligns with the widespread view in philosophy and developmental psychology that abstraction over instances is a main mechanism of concept formation (e.g., the influential work of Jean Piaget).
Let us explain why concepts, and in particular those modeled by distributional semantics (because there is some confusion about their nature), would be suitable representatives of expression meaning.

It is sometimes assumed that the word vector for ``cat'' should model the concept \textsc{Cat} (we discuss some work that makes this assumption below).
This may be a `true enough' approximation for practical applications, but theoretically it is, strictly speaking, on the wrong track.
This is because the word vector for ``cat'' does not model the concept \textsc{Cat} -- that would be an abstraction over occurrences of \emph{actual cats}, after all.
Instead, the word vector for ``cat'' is an abstraction over occurrences of \emph{the word}, not the animal, hence it would model the concept of \emph{the word} ``cat'', say, \textsc{TheWordCat}.
The extralinguistic concept \textsc{Cat} and the linguistic concept \textsc{TheWordCat} are very different.
The concept \textsc{Cat} encodes knowledge about cats having fur, four legs, the tendency to meow, etc.; the concept \textsc{TheWordCat} instead encodes knowledge that the word ``cat'' is a common noun, that it rhymes with ``bat'' and ``hat'', how speakers have used it or tend to use it, that the word doesn't belong to a particular register, and so on.\footnote{
	To clarify: the difference persists even if the notion of context in distributional semantics is enriched to include, say, pictures of cats, or even actual cats.
	The distributions it models would still be distributions \emph{of words}, not of things like cats.}

Our distinction between \textsc{TheWordCat} and \textsc{Cat}, or between linguistic and extralinguistic concepts, is not new, and word vectors are known to capture the more linguistic kind of information, and to be (at best) only a proxy for the extralinguistic concepts they are typically used to denote by a speaker (e.g., \citealt{miller1991contextual}).
But it appears to be sometimes overlooked.
For instance, the assumption that the word vector for ``cat'' would (or should) model the extralinguistic concept \textsc{Cat} is made in work using distributional semantics to model entailment, e.g., that being a cat entails being an animal (e.g., \citealt{geffet2005distributional, roller2014inclusive, vulic2017specialising}).
But clearly the entailment relation holds between the extralinguistic concepts \textsc{Cat} and \textsc{Animal} -- being a cat entails being an animal -- \emph{not} between the linguistic concepts \textsc{TheWordCat} and \textsc{TheWordAnimal} actually modeled by distributional semantics: being the word ``cat'' does not entail (in fact, it excludes) being the word ``animal''.
Hence these approaches are, strictly speaking, theoretically misguided -- although their conflation of linguistic and extralinguistic concepts may be a defensible simplification for practical purposes.

There have been many proposals to integrate formal and distributional semantics (e.g., \citealt{Beltagy:13montague, Erk:13towards, baroni2014frege, asher2016integrating}), and a similar confusion exists in at least some of them~\citep{asher2016integrating,Mcnally:17conceptual}.
We are unable within the scope of the current paper to do justice to the technical sophistication of these approaches, but for present purposes, impressionistically, the type of integration they pursue can be pictured as follows:
\ex. \label{ex:red-cat-integrated} The red cat is chasing a mouse. \\[1pt]
\textbf{Formal semantics:} \ \ \ \ $\iota x (\textsc{Red}(x) \land \textsc{Cat}(x) \land \exists y (\textsc{Mouse}(y) \land \textsc{Chase}(x,y)))$ \\[2pt]
\textbf{Distributional semantics:} \ \ \ $^\nearrow \ \ _\searrow \ \ _\swarrow \ \rightarrow \ \ _\downarrow \ \ ^\nearrow \ \leftarrow$ \ \ \ \ \ \ (i.e., a vector for each word) \\[1pt]
\textbf{Possible integration:} \ \ \ \ $\iota x (_\searrow(x) \land \hspace{.3pt} _\swarrow \hspace{-.5pt} (x) \land \exists y (\ \leftarrow \hspace{-2pt} (y) \land \ _\downarrow \hspace{-.5pt} (x,y)))$ \ \ \ \ (very simplistically)

Again, this may be a `true enough' approximation, but it is theoretically on the wrong track.
The atomic constants in formal semantics are normally understood (e.g., \citealt{frege1892sinn} and basically anywhere since) to denote the extralinguistic kind of concept, i.e., \textsc{Cat} and not \textsc{TheWordCat}.
Put differently, entity $x$ in example \ref{ex:red-cat-integrated} should be entailed to be a cat, not to be the word ``cat''.
This means that the distributional semantic word vectors are, strictly speaking, out of place in a formal semantic skeleton like in~\ref{ex:red-cat-integrated}.\footnote{
	The mathematical techniques of the aforementioned approaches do not depend for their validity on the exact nature of the vectors.
	We hope that these techniques can be used to represent not expression meaning but speaker meaning (see section~\ref{sec:reference}), provided we use vector representations of the distribution of actual cats, instead of the word ``cat''.}

In short, distributional semantics models linguistic concepts like \textsc{TheWordCat}, not extralinguistic concepts like \textsc{Cat}.
But this is not a shortcoming; it makes distributional semantics more adequate, rather than less adequate, as a model of expression meaning, for the following reason.
A prominent strand in the literature on concepts conceives of concepts as \emph{abilities} (e.g., \citealt{dummett1993seas, bennett2008history}; for discussion see \citealt{sep:concepts}).
For instance, possessing the concept \textsc{Cat} amounts to having the ability to recognize cats, discriminate them from non-cats, and draw certain inferences about cats.
The concept \textsc{Cat} is, then, the starting point for \emph{interpreting} an object as a cat and draw inferences from it.
It follows that the concept \textsc{TheWordCat} is the starting point for interpreting a word as the word ``cat'' and drawing inferences from it, notably, inferences about what a speaker in a particular context may use it for: for instance, to refer to a particular cat.\footnote{
	This is because how a speaker may use a word is constrained by how speakers have used it in the past -- a trait of linguistic convention.
	Since the concept \textsc{TheWordCat} reflects uses of ``cat'' in the past, among which are referential uses, it constrains (hence warrants inferences about) what it may be used by a given speaker to refer to.
	(To clarify: this does not imply that the actual or potential referents of a word are actually part of its meaning -- see Section~\ref{sec:reference}.)
	The same holds for the distributional semantic word vector for ``cat'', although instantiations of distributional semantics may differ in how much referentially relevant information they encode.
	Presumably, more information of this sort is encoded when reference is prominent in the original data, for instance when a distributional semantic model is trained on referential expressions grounded in images \citep{KazemzadehOrdonezMattenBergEMNLP14}; otherwise such information needs to be induced from patterns in the text alone (like any other semantic information in text-only distributional semantics).}
Thus, \textbf{the view of distributional semantics as a model of concepts, but crucially concepts \emph{of words}, establishes word vectors as a necessary starting point for interpreting a word.}
This is exactly the explanatory job assigned to expression meaning: a context-invariant starting point for interpretation.
Not coincidentally, for neural networks that take words as input, distributional semantics resides in the first layer of weights (see Section~\ref{sec:whatwemean}).

Summing up, this section presented two reasons why distributional semantics is attractive as a model of expression meaning. The next section considers whether it could also be \emph{sufficient}.

\section{Limits of distributional semantics: words don't refer, speakers do.}
\label{sec:reference}

In many ways the standard for what a theory of expression meaning ought to do has been set by formal semantics. 
Consider again our simplistic comparison of distributional semantics and formal semantics:
\ex. \label{ex:red-cat-formal2} The red cat is chasing a mouse. \\[1pt]
\textbf{Formal semantics:} \ \ \ \ $\iota x (\textsc{Red}(x) \land \textsc{Cat}(x) \land \exists y (\textsc{Mouse}(y) \land \textsc{Chase}(x,y)))$ \\[2pt]
\textbf{Distributional semantics:} \ \ \ $^\nearrow \ \ _\searrow \ \ _\swarrow \ \rightarrow \ \ _\downarrow \ \ ^\nearrow \ \leftarrow$ \ \ \ \ \ \ (i.e., a vector for each word)

The logical formulae into which formal semantics translates this example are assigned precise interpretations in (a model of) the outside world.
For instance, \textsc{Red} would denote the set of all red things, \textsc{Cat} the set of all cat-like things, \textsc{Chase} a set of pairs where one chases the other, the variable $x$ would be bound to a particular entity in the world, etc., and the logical connectives can have their usual truth-conditional interpretation.\footnote{
	In fact, the common reliance on an intermediate formal, logical language is not what defines formal semantics; what matters is that it treats natural language itself as a formal language \citep{Montague:70universal}, by compositionally assigning precise interpretations to it -- and this can be done directly, or indirectly via translation to a logical language as in our example.}
In this way formal semantics accounts for reference to things in the world and it accounts for truth values (which is what sentences refer to; \citealt{frege1892sinn}).
Moreover, referents and truth values across possible worlds/situations in turn determine truth conditions, and thereby entailments -- because one sentence entails another if whenever the former is true the latter is true as well.\footnote{
	There are serious shortcomings to the formal semantics approach, some of which we discuss below, but others which aren't relevant for present purposes.
	An important criticism that we won't discuss is that the way in which formal semantics assigns interpretations to natural language relies crucially on the manual labor of hard-working semanticists, which does not scale up.
	}
By contrast, distributional semantics on its own (cf.\ footnote~\ref{fn:onitsown}) struggles with these aspects (\citealt{BoledaHerbelot:16}; see also the work discussed in section~\ref{sec:concepts} on entailment), which has motivated aforementioned attempts to integrate formal and distributional semantics (e.g., \citealt{Beltagy:13montague, Erk:13towards, baroni2014frege, asher2016integrating, BoledaHerbelot:16}).
Put simply, distributional semantics struggles because there are no entities or truth values in distributional space to refer to.
Nevertheless, we think that this isn't a shortcoming of distributional semantics; \textbf{we argue that a theory of expression meaning \emph{shouldn't} model these aspects}.\footnote{
	Truth conditions, entailments and reference are just three sides of the same central, referential tenet of formal semantics, and what we will say about reference in what follows will apply to truth conditions and entailment, and vice versa. 
	An anonymous reviewer draws our attention also to the logical notions of satisfiability and validity, i.e., possible vs.\ necessary truth.
	Our proposal applies to these notions too, regardless of whether they are understood in terms of quantification over possible ways the world may be, or in terms of quantification over possible interpretations.}

We think that these referential notions on which formal semantics has focused are best understood to reside at the level of speaker meaning, not expression meaning.
In a nutshell, \textbf{our position is that words don't refer, speakers do} (e.g., \citealt{Strawson:50referring}) -- and analogously for truth conditions and entailment.
The fact that speakers often refer \emph{by means of} linguistic expressions doesn't entail that these expressions must in themselves, out of context, have a determinate reference, or even be capable of referring (or capable of entailing, of providing information, of being true or false).
Parsimony (again) suggests that we do not assume the latter:
To explain why a speaker can use, e.g., the expression ``cat'' to refer to a cat, it is sufficient that, in the relevant community, that is how the expression is often used.
It is theoretically superfluous to assume in addition that the expression ``cat'' itself refers to cats.

Now, most work in formal semantics would acknowledge that ``cat'' out of context doesn't refer to cats, and that its use in a particular context to refer to cats must be explained on the basis of a less determinate, more underspecified notion of expression meaning.
More generally, expressions are well-known to underdetermine speaker meaning (e.g., \citealt{Bach:94, Recanati:04}), as basically any example can illustrate (e.g.,~\ref{ex:red-cat} ``red cat'' and~\ref{ex:marry1} ``got married'').
However, this alone does not imply that the notions of formal semantics are inadequate for characterizing expression meaning; in principle one could try to define, in formal semantics, the referential potential of ``cat'' in a way that is compatible with its use to refer to cats, to cat-like things, etcetera.
And one could define the expression meaning of ``Jack and Jill got married'' in a way that is compatible with them marrying each other and with each marrying someone else.\footnote{
	For instance, an anonymous reviewer notes that richer logical formalisms such as dependent type theory are well-suited for integrating contextual information into symbolic representations.
}
What is problematic for a formal semantic approach is that the \emph{ways} in which expressions underdetermine speaker meaning are not clearly delineated and enumerable, and that there is no symbolically definable common core among all uses.\footnote{
	Similarly, Bach (\citeyear{bach2005context}, among others) has criticized the common approach in formal semantics of incorporating, in definitions of expression meaning, `slots' where supposed context-sensitive material is to be plugged in.
	The meaning of a scalar adjective like ``big'', for instance, would contain a slot for `standard of comparison' to be filled by context in order to explain why the same thing may be described as ``big'' in one context but not in another (e.g., \citealt{kennedy2007vagueness}).
	\citet{bach2005context} notes that this type of approach does not generalize to all the ways in which expression meaning underdetermines speaker meaning; the meaning of each expression would essentially end up being a big empty slot, to be magically filled by context.
}
This argument was made for instance by \citet{Wittgenstein:53}, who notes that the uses of an expression (his example was ``game'') are tied together not by definition but by family resemblance.
More recent iterations of this argument can be found in criticisms of the ``classical'', definitional view of concepts (e.g., \citealt{rosch1975family, fodor1980against, sep:concepts}),
\notiwcs{It is also demonstrated by the poor track record of the definitional approach: as noted by \citet{sep:concepts}, no single definition of a word or concept has in philosophy ever commanded general agreement, not even a term as intensively studied as, e.g., ``knowledge''.}
and in criticisms of sense enumeration approaches to word meaning (e.g., \citealt{kilgarriff1997don, hanks2000word, erk2010word}; cf.\ \citealt{pustejovsky1995}), which we already mentioned briefly before: it is unclear what constitutes a word sense, and no enumeration of senses covers all uses.

\textbf{The only truly common core among all uses of any given expression is that they are all, indeed, uses of the same expression.}
Hence, if expression meaning is to serve its purpose as a common core among all uses, i.e., as a context-invariant starting point of semantic/pragmatic explanations, then it must reflect all uses.
As we argued in section~\ref{sec:ds}, distributional semantics, conceived of as a model of expression meaning (i.e., the `strong' view of \citealt{Lenci:08}), embraces exactly this fact.
This makes the representations of distributional semantics, but not those of formal semantics, suitable for characterizing expression meaning.
By contrast, (largely) discrete notions like reference, truth and entailment are useful, at best, at the level of \emph{speaker} meaning -- recall that our position is that words don't refer, speakers do \citep{Strawson:50referring}.\footnote{
	We are not discussing another long-standing criticism of formal semantics, namely that referring (and asserting something that can be true or false) is not all that speakers do with language (e.g., \citealt{austin1975things, searle1969speech}). We do not claim that formal semantics would be \emph{sufficient} as a model of speaker meaning; only that its notions are more adequate there than at the level of expression meaning.
}
That is, one can fruitfully conceive of a particular speaker, in some individuated context, as intending to refer to discrete things, communicating a certain determinate piece of information that can be true or false, entailing certain things and not others.
This still involves considerable abstraction, as any symbolic model of a cognitive system would \citep{Marr:83}; e.g., speaker intentions may not always be as determinate as a symbolic model presupposes.
But the amount of abstraction required, in particular the kind of determinacy of content that a symbolic model presupposes, is not as problematic in the case of speaker meaning as for expression meaning.
The reason is that a model of speaker meaning needs to cover only a single usage, by a particular speaker situated in a particular context; a model of expression meaning, by contrast, needs to cover countless interactions, across many different contexts, of a whole community of speakers.
The symbolic representations of formal semantics are ill-suited for the latter.

Despite the foregoing considerations being prominent in the literature, formal semantics has continued to assume that referents, truth conditions, etc., are core aspects of expression meaning.
The main reason for this is the traditional centrality of supposedly `semantic' intuitions in formal semantics \citep{Bach:02}, either as the main source of data or as the object of investigation (`semantic competence', for criticism see \citealt{Stokhof:11}).
In particular, formal semantics has attached great importance to intuitions about truth conditions (e.g., ``semantics with no treatment of truth conditions is not semantics'', \citealt{Lewis:72}:169), a tenet going back to its roots in formal logic (e.g., \citealt{Montague:70universal} and the earlier work of Frege, Tarski, among others).
Clearly, if expressions on their own do not even \emph{have} truth conditions, as we have argued, these supposedly semantic intuitions cannot genuinely be about expression meaning.
And that is indeed what many authors have pointed out. 
\citet{Strawson:50referring, Grice:75, Bach:02}, among others, have argued that \textbf{what seem to be intuitions about the meaning of an expression are really about what a stereotypical speaker would mean by it} -- or at least they are heavily influenced by it.
Again example~\ref{ex:marry1} serves as an illustration here: intuitively ``marry'' means ``marry each other'', but to assume that this is therefore its expression meaning would be inadequate (as we discussed in section~\ref{sec:meaningfromuse}).
But we want to stress that this is not just an occasional trap set by particular kinds of examples; just being a bit more careful doesn't cut it.
It is the foundational intuition that expressions can even \emph{have} truth conditions that is already inaccurate.
Our intuitions are \emph{fundamentally} not attuned to expression meaning, because expression meaning is not normally what matters to us; it is only an instrument for conveying speaker meaning, and, much like the way we string phonemes together to form words, it plays this role largely or entirely without our conscious awareness.
The same point has been made in the more psycholinguistic literature \citep{Schwarz:96}, occasionally in the formal semantics/pragmatics literature \citep{KadmonRoberts:86}, and there is increasing acknowledgment of this also in experimental pragmatics, in particular of the fact that participants in experiments imagine stereotypical contexts (e.g., \citealt{WesteraBrasoveanu:14salt, degen2015processing, poortman2017concept}).

Summing up, the standard that formal semantics has set for what a theory of expression meaning ought to account for, and which makes distributional semantics appear to fall short, turns out to be misguided.
Reference, truth conditions and entailment belong at the level of speaker meaning, not expression meaning.
It entails that distributional semantics on its own need not account for these aspects, either theoretically or computationally; it should only provide an adequate starting point.
Interestingly, this corresponds exactly to its role in current neural network models, on tasks that involve identifying aspects of speaker meaning.
Consider the task of visual reference resolution (e.g., \citealt{plummer2015flickr30k}), where the inputs are a linguistic description plus an image and the task is to identify the intended referent in the image.
A typical neural network model would achieve this by first activating word embeddings (a form of distributional semantics; Section~\ref{sec:whatwemean}) and then combining and transforming these together with a representation of the image into a representation of the intended referent -- speaker meaning.

\section{Compositionality}
\label{sec:compositionality}

Language is compositional in the sense that what a larger, composite expression means is determined (in large part) by what its components mean and the way they are put together. 
Compositionality is sometimes mentioned as a strength of formal semantics and as an area where distributional semantics falls short~\citep[a.o.][]{Beltagy:13montague}.
But in fact both approaches have shown strengths and weaknesses regarding compositionality (see \citealt{BoledaHerbelot:16} for an overview).
To illustrate, consider again:
\ex. \label{ex:red-cat9} The red cat is chasing a mouse.

In this context the adjective ``red'' is used by the speaker to mean something closer to \textsc{Orange} (because the ``red hair'' of cats is typically orange), unlike its occurrence in, say, ``red paint''.
Distributional semantics works quite well for this type of effect in the composition of content words (e.g., \citealt{baroni2014frege, Mcnally:17conceptual}), an area where formal semantics, which tends to leave the basic concepts unanalyzed, has struggled (despite efforts such as \citealt{pustejovsky1995}).
Classic compositional distributional semantics, in which distributional representations are combined with some externally specified algorithm (which can be as simple as addition), also works reasonably well for short sentences, as measured for instance on sentence similarity (e.g., \citealt{mitchell2010composition,Grefenstette2013,marelli2014sick}).
But for longer expressions distributional semantics on its own falls short (cf.\ our clarification of ``on its own'' in footnote~\ref{fn:onitsown}), and this is part of what has inspired aforementioned works on integrating formal and distributional semantics (e.g., \citealt{coecke2011mathematical,Grefenstette2011,Beltagy:13montague, Erk:13towards, baroni2014frege, asher2016integrating}).

However, that distributional semantics falls short of accounting for full-fledged compositionality does not mean that it cannot be a sufficient model of expression meaning.
For that, it should be established first that compositionality wholly resides at the level of expression meaning -- and it is not clear that it does.
Let us take a closer look at the main theoretical argument for compositionality, the argument from \emph{productivity}.\footnote{
	To clarify: the issue here is not whether \emph{distributed} representations can be composed, but whether \emph{distributional} representations -- i.e., abstractions over distributions of use -- can and should be composed. 
	Sophisticated approaches exist for composing \emph{distributed} representations (notably the tensor product approach of \citealt{smolensky1990tensor}).}
According to this argument, compositionality is necessary to explain how a competent speaker can understand the meaning of a composite expression that they have never before encountered.
However, in appealing to a person's supposed understanding of the meaning of an expression, this argument is subject to the revision proposed in Section~\ref{sec:reference}: 
it reflects speaker meaning, not expression meaning.
More correctly phrased, then, the type of data motivating the productivity argument is that a person who has never encountered a speaker uttering a certain composite expression, is nevertheless able to understand \emph{what some (actual or hypothetical) speaker would mean by it}.
And this leaves undetermined where compositionality should reside: at the level of expression meaning, speaker meaning, or both.

To illustrate, consider again example~\ref{ex:red-cat9}, ``The red cat is chasing a mouse''.
A speaker of English who has never encountered this sentence will nevertheless understand what a stereotypical speaker would mean by it (or will come up with a set of interpretations) -- this is an instance of productivity.
One explanation for this would be that the person can compositionally compute an expression meaning for the whole sentence, and from there infer what a speaker would mean by it.
This places the burden of compositionality entirely on the notion of expression meaning.
An alternative would be to say that the person first infers speaker meanings for each word (say, the concept \textsc{Cat} for ``cat''),\footnote{
	We discuss this here as a hypothetical possibility; to assume that individual words of an utterance can be assigned speaker meanings may not be a feasible approach in general.
} and then composes these to obtain a speaker meaning of the full sentence.
This would place the burden of compositionality entirely on the notion of speaker meaning (cf.\ the notion of \emph{resultant procedure} in \citealt{grice1968utterer}; see \citealt{borge2009intentions} for a philosophical argument for compositionality residing at the speaker meaning level).
The two alternatives are opposite extremes of a spectrum; and note that the first is what formal semantics proclaims, yet the second is what formal semantics does, given that the notions it composes in fact reside at the level of speaker meaning (e.g., concepts like \textsc{Cat} as opposed to \textsc{TheWordCat}; and the end product of composition in formal semantics is typically a truth value).
There is also a middle way:
The person could in principle compositionally compute expression meanings for certain intermediate constituents (say, ``the red cat'', ``a mouse'' and ``chases''), then infer speaker meanings for these constituents (say, a particular cat, an unknown mouse, and a chasing event), and only \emph{then} continue to compose these to obtain a speaker meaning for the whole sentence.
This kind of middle way requires that a model of expression meaning (distributional semantics) accounts for some degree of compositionality (say, the direct combination of content words), with a model of speaker meaning (say, formal semantics) carrying the rest of the burden.
The proposal in~\citet{Mcnally:17conceptual} is a version of this position.

The foregoing shows that the productivity argument for compositionality falls short as an argument for compositionality \emph{of expression meanings}; that is, \textbf{compositionality may well reside in part, or even entirely, at the level of speaker meaning}.
We will not at present try to settle the issue of where compositionality resides -- though we favor a view according to which compositionality is multi-faceted and doesn't necessarily reside exclusively at one level.%
\footnote{
	The empirical picture is undecisive in this regard: just because distributional semantics appears to be able to handle certain aspects of compositionality, that doesn't mean it should.
	After all, word vectors like ``cat'' have been quite successfully used as a proxy for extra-linguistic concepts like \textsc{Cat}, even though as we explained this is strictly speaking a misuse (conflating \textsc{Cat} and \textsc{TheWordCat}; see section~\ref{sec:concepts}).
	Perhaps the moderate success of distributional semantics on for instance adjective-noun composition like ``red cat'' reflects the fact that the extra-linguistic concepts \textsc{Red} and \textsc{Cat} compose (speaker meaning), even if the linguistic concepts \textsc{TheWordRed} and \textsc{TheWordCat} don't (expression meaning).}
What matters for the purposes of this paper is that the requirement imposed by formal semantics, that a theory of expression meaning should account for full-fledged compositionality, turns out to be unjustified.

\section{Outlook}

We presented two strong reasons why distributional semantics is attractive as a model of expression meaning, i.e., in favor of the `strong' view of \citealt{Lenci:08}:
The parsimony of regarding expression meaning as an abstraction over use; and the understanding of these abstractions as concepts and, thereby, as a necessary starting point for interpretation.
Moreover, although distributional semantics struggles with matters like reference, truth conditions and entailment, we argued that a theory of expression meaning \emph{should not} account for these aspects: words don't refer, speakers do (and likewise for truth conditions and entailments).
The referential approach to expression meaning of formal semantics is based on misinterpreting intuitions about stereotypical speaker meaning as being about expression meaning.
The same misinterpretation has led to the common view that a theory of expression meaning should be compositional, whereas in fact compositionality may reside wholly or in part (and does reside, in formal semantics) at the level of speaker meaning.
Clearing this up reveals that distributional semantics is the more adequate approach to expression meaning.
In between our mostly theoretical arguments for this position, we have shown how a consistent interpretation of distributional semantics as a model of expression meaning sheds new light on certain applications: e.g., distributional semantic approaches to entailment and attempts at integrating distributional and formal semantics.

\section*{Acknowledgments}
We are grateful to the anonymous reviewers for their valuable comments. This project has received funding from the European Research Council (ERC) under the European Union's Horizon 2020 research and innovation programme (grant agreement No 715154), and from the Spanish Ram\'on y Cajal programme (grant RYC-2015-18907). This paper reflects the authors' view only, and the EU is not responsible for any use that may be made of the information it contains.
\begin{flushright}
	\includegraphics[width=0.8cm]{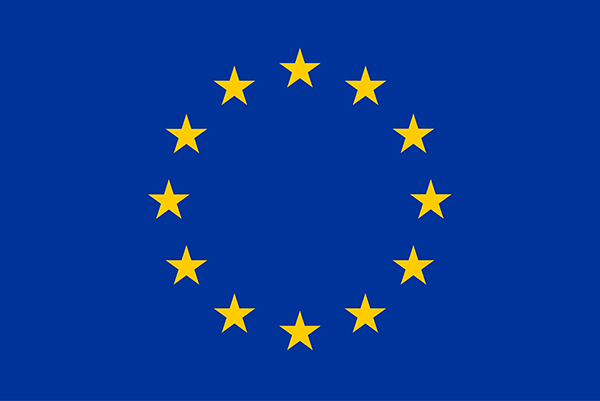}  
	\includegraphics[width=0.8cm]{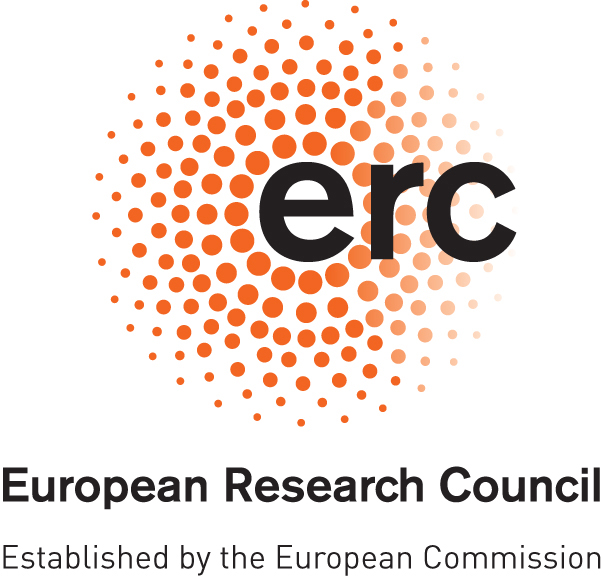} 
\end{flushright}

\bibliographystyle{chicago}
\bibliography{formaldistributional,gemma}

\end{document}